\documentclass[letterpaper, 10 pt, conference]{ieeeconf}  

\IEEEoverridecommandlockouts                              

\usepackage{graphicx, subcaption, booktabs, multirow, makecell,wrapfig}
\usepackage{dblfloatfix}
\usepackage{amsmath} 
\usepackage{amssymb}  
\usepackage{array}
\usepackage{tabularx}
\usepackage[table,xcdraw]{xcolor}

\title{\LARGE \bf
RoboCSE: Robot Common Sense Embedding
}

\author{Angel Daruna, Weiyu Liu, Zsolt Kira, Sonia Chernova
\thanks{Institute for Robotics and Intelligent Machines. 
  Georgia Institute of Technology, Atlanta, Georgia. United States
        {\tt\small adaruna3,wliu88,zkira@gatech.edu;\newline chernova@cc.gatech.edu} }%
}

\begin{document}

\newcolumntype{L}[1]{>{\raggedright\arraybackslash}p{#1}}
\newcolumntype{C}[1]{>{\centering\arraybackslash}p{#1}}
\newcolumntype{R}[1]{>{\raggedleft\arraybackslash}p{#1}}

\maketitle
\thispagestyle{empty}
\pagestyle{empty}

\begin{abstract}

Autonomous service robots require computational frameworks that allow them to generalize knowledge to new situations in a manner that models uncertainty while scaling to real-world problem sizes.  The Robot Common Sense Embedding (RoboCSE) showcases a class of computational frameworks, multi-relational embeddings, that have not been leveraged in robotics to model semantic knowledge.  We validate RoboCSE on a realistic home environment simulator (AI2Thor) to measure how well it generalizes learned knowledge about object affordances, locations, and materials. Our experiments show that RoboCSE can perform prediction better than a baseline that uses pre-trained embeddings, such as Word2Vec, achieving statistically significant improvements while using orders of magnitude less memory than our Bayesian Logic Network baseline. In addition, we show that predictions made by RoboCSE are robust to significant reductions in data available for training as well as domain transfer to MatterPort3D, achieving statistically significant improvements over a baseline that memorizes training data.
	

\end{abstract}



\section{Introduction}

Robots operating in human environments benefit from encoding world information in a semantically-meaningful representation in order to facilitate generalization and domain transfer.  This work focuses on the problem of semantically representing a robot's world in a robust, generalizable, and scalable fashion.  Semantic knowledge is typically modeled by a set of entities $\mathcal{E}$ representing concepts known to the robot (e.g. apple, fabric, kitchen), and a set of possible relations $\mathcal{R}$ (e.g. atLocation, hasMaterial, hasAffordance) between them \cite{tenorth2013knowrob,chernovasituated,saxena2014robobrain,zhu2014reasoning,celikkanat2015probabilistic,beetz2018know}.

While some semantic information can be hard-coded, large-scale and long-term deployments of autonomous systems require the development of computational frameworks that i) enable abstract concepts to be learned and generalized from observations, ii) effectively model the uncertain nature of complex real-world environment, and iii) are scalable, incorporating data from a wide range of environments (e.g., hundreds of households).  Previous work in semantic reasoning for robot systems has addressed subsets of the above challenges.  Directed graphs \cite{koller2009probabilistic} used in \cite{saxena2014robobrain} allowed individual observations to adapt generalized concepts at large scale, integrating multiple projects.  Bayesian Logic Networks (BLN) \cite{jain2009bayesian} in \cite{chernovasituated} allowed for precise probabilistic inference and learning assuming knowledge graphs have manageable sizes.  Description Logics (DL) \cite{baader2008description} used in \cite{tenorth2010knowrob} allowed for large-scale deterministic reasoning about many concepts. In summary, each of these representations have limitations with respect to at least one of the three characteristics above.
    
\begin{figure}
        \centering
        \includegraphics[width=0.8\columnwidth]{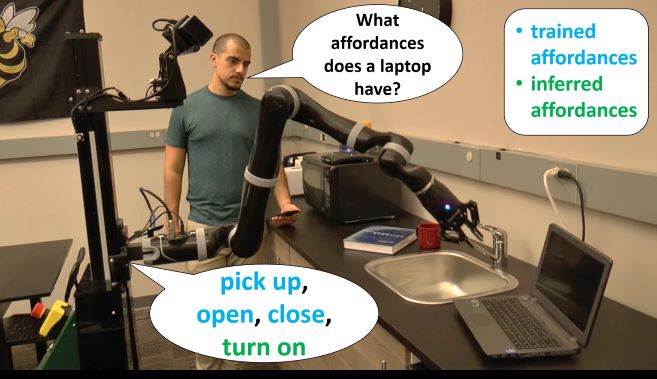}
        \caption[]
        {\small RoboCSE can be queried by a robot to infer knowledge and make decisions.} 
        \label{fig:robot_demo}
        \vspace{-.6cm}
\end{figure}

\begin{figure*}
        \centering
        \begin{subfigure}[b]{0.25\columnwidth}
            \centering
            \includegraphics[width=\columnwidth]{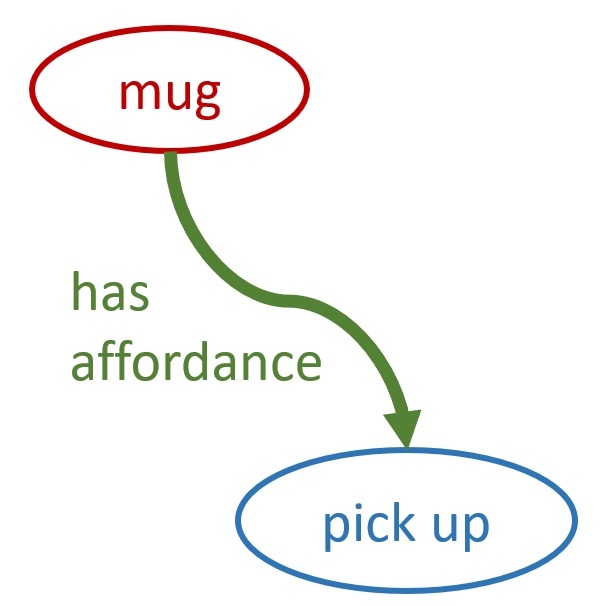}
            \caption[]%
            {{}}    
            \label{fig:dg}
        \end{subfigure}
        \quad
        \begin{subfigure}[b]{0.25\columnwidth}
            \centering
            \includegraphics[width=\columnwidth]{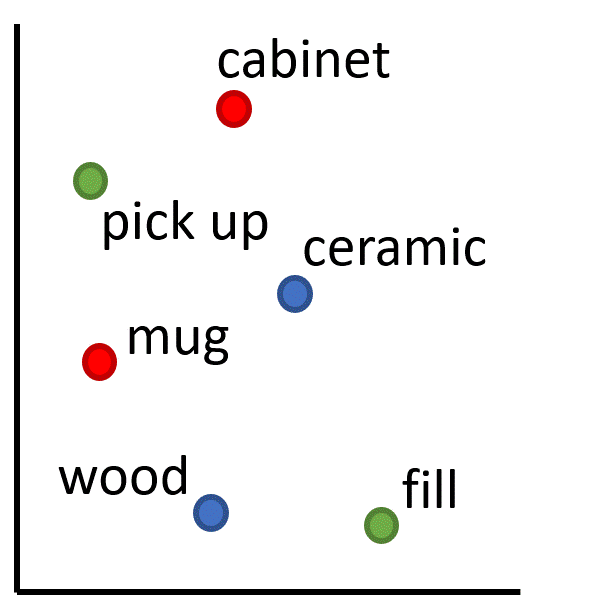}
            \caption[]%
            {{}}    
            \label{fig:random_init}
        \end{subfigure}\quad
        \begin{subfigure}[b]{0.28\columnwidth}   
            \centering 
            \includegraphics[width=\columnwidth]{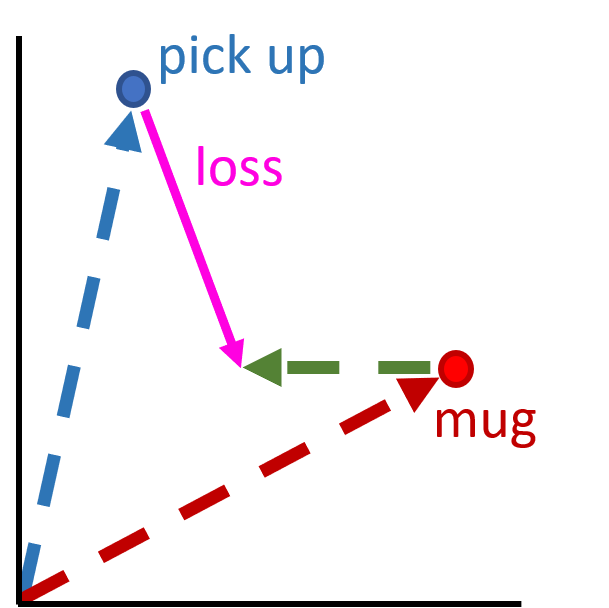}
            \caption[]%
            {{}}    
            \label{fig:pre_update}
        \end{subfigure}\quad
        \begin{subfigure}[b]{0.28\columnwidth}
            \centering 
            \includegraphics[width=\columnwidth]{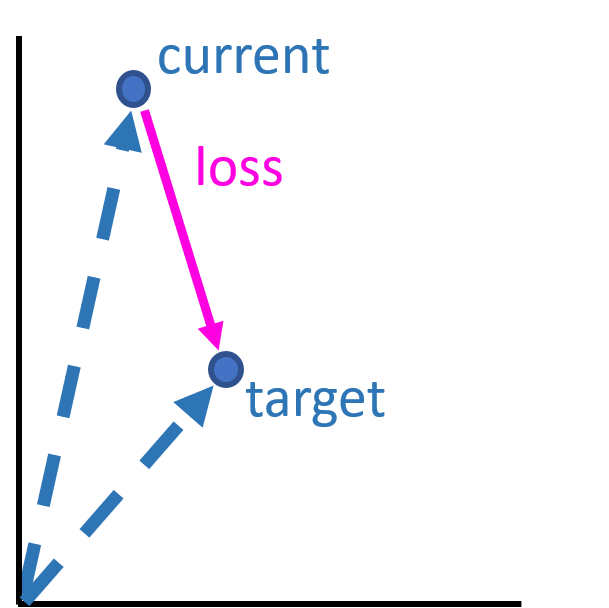}
            \caption[]%
            {{}}    
            \label{fig:update}
        \end{subfigure}\quad
        \begin{subfigure}[b]{0.28\columnwidth}   
            \centering 
            \includegraphics[width=\columnwidth]{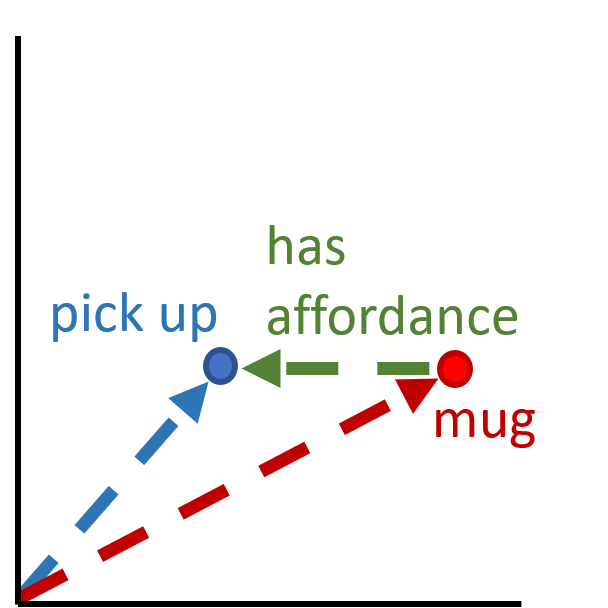}
            \caption[]%
            {{}}    
            \label{fig:post_update}
        \end{subfigure}\quad
        \begin{subfigure}[b]{0.25\columnwidth}  
            \centering 
            \includegraphics[width=\columnwidth]{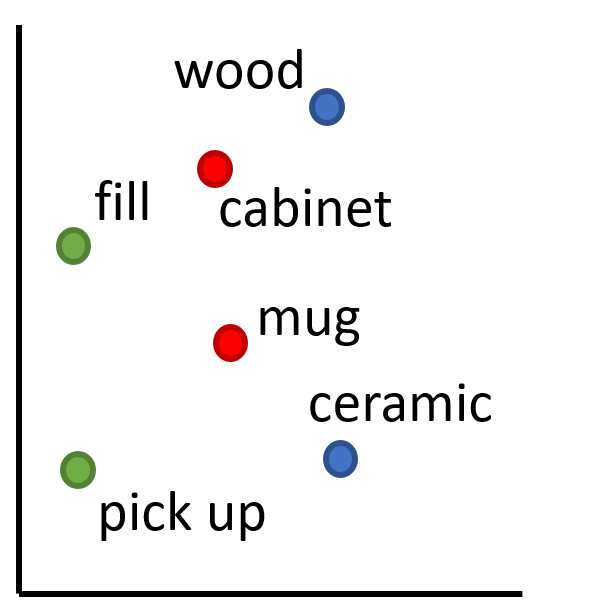}
            \caption[]%
            {{}}    
            \label{fig:learned}
        \end{subfigure}
        \caption[]
        {\small From a directed graph (\ref{fig:dg}) we can learn a vector embedding containing the same nodes and edges. Multi-relational embeddings begin randomly initialized (\ref{fig:random_init}) and are updated by calculating the losses between target transformations and actual transformations (\ref{fig:pre_update}-\ref{fig:update}) until they converge on a semantically meaningful structure (\ref{fig:post_update}-\ref{fig:learned}).}
        \label{fig:learning}
        \vspace{-.6cm}
\end{figure*}

In this work, we contribute Robot Common Sense Embedding (RoboCSE), a novel computational framework for semantic reasoning that is highly scalable, robust to uncertainty, and generalizes learned semantics.  Given a knowledge graph $\mathcal{G}$, formalized in Section~\ref{sec:background}, RoboCSE encodes semantic knowledge using \textit{multi-relational embeddings} \cite{nickel2016review}, embedding $\mathcal{G}$ into a high-dimensional vector space that preserves graphical structure between nodes and edges, while also facilitating generalization (see Figure~\ref{fig:dg},\ref{fig:post_update}). We show that RoboCSE can be trained on simulated environments (AI2Thor \cite{kolve2017ai2}), and that the resulting learned model effectively transfers to data from real-world domains, including both pre-recorded household scenes (MatterPort3D \cite{chang2017matterport3d}) and real-time execution on a robot\footnote{https://youtu.be/ynHwNotCkDA} (Figure~\ref{fig:robot_demo}). 

We compare our work to three baselines: BLNs, Google's pre-trained Word2Vec embeddings \cite{mikolov2013distributed}, and a theoretical upper bound on the performance of logic-based methods \cite{baader2008description}. Our results show that RoboCSE uses orders of magnitude less memory than BLNs\footnote{implemented using ProbCog \cite{jain2010soft}} and outperforms the Word2Vec and logic-based baselines across all accuracy metrics.  RoboCSE also successfully generalizes beyond the training data, inferring triples held-out from the training set, by leveraging latent interactions between multiple relations for a given entity.  Furthermore, results returned by RoboCSE are ranked by confidence score, enabling robot behavior architectures to be constructed that effectively reason about the level of uncertainty in the robot's knowledge.  Combined, the memory efficiency and learned generalizations of RoboCSE allow a robot to semantically model a larger world while accounting for uncertainty.

\section{Related Work}
\label{sec:relateds}
    
Data-driven methods using convolutional neural networks have shown promise in semantically reasoning about high-level tasks \cite{zhu2017visual} and trajectory selection \cite{sung2018robobarista}. However, the representations learned by these methods can lose the graph structure beneficial for reasoning, often require large amounts of data, and use end-to-end pipelines that could lead to performance drops across domains even with similar semantics. Inspired by data-driven approaches, RoboCSE is a learned representation but makes simplifying mathematical assumptions about the embedding space to promote structure.  In addition, we decouple perceptual stimuli from semantics by operating on symbols (vectors) to allow domain transfer without tuning (e.g. virtual vs. real-world).

We propose to use multi-relational embeddings to learn a knowledge graph $\mathcal{G}$. Multi-relational embeddings represent knowledge graphs in vector space, encoding vertices that represent entities $\mathcal{E}$ as vectors and edges that represent relations $\mathcal{R}$ as mappings. The simplest of these models are \textit{Translational Methods} \cite{wang2017kge_survey} such as TransE \cite{bordes2013transe}, TransH \cite{wang2014transh}, and TransR \cite{lin2015transr}. \textit{Semantic Matching Methods} have outperformed translational methods because they offer a  wider range of possible relations between entities than vector addition \cite{wang2017kge_survey}. Semantic matching methods \cite{socher2013reasoning,dong2014knowledge} leverage neural-networks to offer relations that can capture non-linear transforms between entities. However, the increase in modeling parameters requires more data and training to avoid over-fitting, which may be difficult for robot systems to acquire.

Our work uses ANALOGY \cite{liu2017analogical}, a semantic matching method, to learn multi-relational embeddings. ANALOGY constrains relations to be normal linear mappings between entities to promote structure in the learned embeddings and simplify the optimization objective. The multiplicative relationship between entities allows for more complex relations to be expressed than vector addition while only requiring a single matrix per relation, balancing scalability with expressiveness to achieve state-of-the-art results \cite{wang2017kge_survey}. 
\section{Approach}
\label{sec:approach}

\subsection{Background: Multi-Relational Embeddings}
\label{sec:background}

The objective of the multi-relational (i.e. knowledge graph) embedding problem is to learn a continuous vector representation of a knowledge graph $\mathcal{G}$, encoding vertices that represent entities $\mathcal{E}$ as a set of vectors $v_\mathcal{E} \in \mathbb{R}^{|\mathcal{E}| \times d_\mathcal{E}}$ and edges that represent relations $\mathcal{R}$ as mappings between vectors $W_\mathcal{R} \in \mathbb{R}^{|\mathcal{R}| \times d_\mathcal{R}}$, where $d_\mathcal{E}$ and $d_\mathcal{R}$ are the dimensions of $\mathcal{E}$ vectors and $\mathcal{R}$ mappings, respectively \cite{nickel2016review,wang2017kge_survey}. The knowledge graph $\mathcal{G}$ is composed from individual knowledge triples $(h,r,t)$ such that $h,t \in \mathcal{E}$ are identified as head and tail entities of the triple, respectively, for which the relation $r \in \mathcal{R}$ holds (e.g. $($\textit{cup, hasAffordance, fill}$)$). Collectively, the set of all triples from a dataset $\mathcal{D}$ form a directed graph $\mathcal{G}$ expressing the knowledge for that domain (note this directed graph is considered incomplete because some set of triples may be missing).

Generically, a multi-relational embedding is learned by minimizing the loss $\mathcal{L}$ using a scoring function $f(h,r,t)$ over the set of knowledge triples from $\mathcal{G}$. In addition to knowledge triples from $\mathcal{G}$, embedding performance substantially improves when negative triples are sampled from a negative triple knowledge graph $\hat{\mathcal{G}}$ \cite{nickel2016review}. Therefore, $\mathcal{L}$ is defined as $\mathcal{L}\big(f(h,r,t),y\big)$ where y is the positive or negative label for the triple.  


\subsection{RoboCSE}
\label{sec:details}

\begin{figure}
      \centering                    
      \includegraphics[width=0.8\columnwidth]{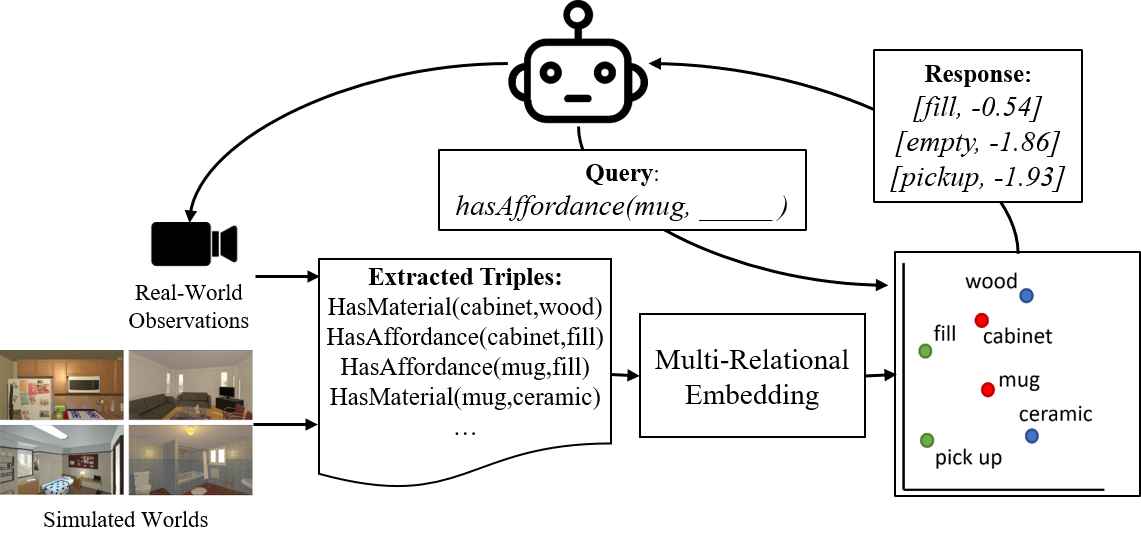}
      \caption[]
      {\small Overview of RoboCSE framework integrated with mobile robot.}
      \label{fig:system_diagram}
      \vspace{-0.5cm}
\end{figure}

RoboCSE is a computational framework for semantic reasoning that uses multi-relational embeddings to encode abstract knowledge obtained by the robot from its sensors, simulation, or even external knowledge graphs (Figure~\ref{fig:system_diagram}).  The robot can use the resulting knowledge representation as a queriable database to obtain information about its environment, such as likely object locations, material properties of objects, object affordances, and any other relation-based semantic information the robot is able to mine.

Figure~\ref{fig:learning} summarizes the embedding process, conceptually.  Training instances are provided in the form of knowledge triples $(h,r,t,y)$ (Figure~\ref{fig:dg}).  The embedding space containing all entity vectors has no structure before training (Figure~\ref{fig:random_init}) because the relational embeddings must be learned.  Therefore, all entities $\mathcal{E}$ and relations $\mathcal{R}$ are initialized as random vectors and mappings, respectively.
    
Each $(h,r,t)$ training instance provided is used to perform stochastic-gradient-descent.  The loss function is formulated as $\mathcal{L}\big(f(h,r,t),y\big) = -\log \sigma (y \cdot f(h,r,t))$, where $\sigma$ is the sigmoid function and $f$ is a bilinear scoring function of the form $f(h,r,t) = \langle v^T_h W_r$,$ v_t \rangle$ \cite{liu2017analogical}. Given a particular multi-relational embedding, its loss function is used to compute a loss between a current vector and a target vector (Figure~\ref{fig:update}). The current vector is calculated using a subset of the knowledge triple (e.g. \textit{pick up} in Figure~\ref{fig:pre_update}) and the target vector is calculated using the remaining subset (e.g. \textit{mug hasAffordance} in Figure~\ref{fig:pre_update}). This loss, is used to update the appropriate $\mathcal{E}$ vectors and $\mathcal{R}$ mappings to better approximate the correct representation (Figure~\ref{fig:post_update}).

The vectors and mappings of $\mathcal{E}$ and $\mathcal{R}$, respectively, converge to semantically meaningful values after repeating the training process with different subsets of knowledge instances (Figure~\ref{fig:post_update}), which can be used to perform inference (see \cite{nickel2016review} for more on learning in multi-relational embeddings). In Figure~\ref{fig:learned} we see that similar entities are grouped horizontally, cabinets are more likely to be filled than picked up, and mugs are equally likely to have either affordance.

Inference in RoboCSE is done by completing a knowledge triple given only partial information. For example given $(h,r\line(1,0){5})$, RoboCSE returns a list of the most likely tails $t_i$ to complete the knowledge instance. Mathematically, given $(h,r,\line(1,0){5})$, $r$ maps an $h$ by some transformation, then the vectors with the highest $f$ scores to the resultant are selected as results, which represent the most likely tails $t_i$. In the case of RoboCSE, which uses \cite{liu2017analogical}, $r$ maps an $h$ via $v^T_h W_r$. Result tails $t_i$ are ordered using the bilinear scoring function $f$, in which higher scores be more likely (i.e. more closely aligned vectors). RoboCSE can make inferences about knowledge triples it has never seen before because these transformations can be done over any entities in the embedding space, allowing for generalization.

An assumption made widely across prior multi-relational embedding work \cite{bordes2013transe,liu2017analogical,wang2017kge_survey} is that query responses are deterministic (i.e. either always true with rank 1 or false with lower ranks), and only factual relations are provided in the training data. However, the semantic data we are modeling is highly stochastic; for example, multiple potential locations are likely for a given object.  As a result, the ground truth rank of responses is often not 1. Instead, ground truth ranks reflect the number of observations in the data.  To support this in our evaluation, we extend the standard performance metrics of mean-reciprocal-rank (MRR) and hits-at-top-k (Hits@K), which assume a ground truth rank of 1, and for the experiments in Section~\ref{sec:exp_res}, we instead report:
\vspace{-0.2cm}
\begin{equation}
	\begin{aligned}
		\small{\textrm{Hits@5*} = \frac{1}{N}\sum_{n=1}^{N} Hits_5\big(\mid G_r-I_r \mid\big)} \\
		\small{s.t. \quad Hits_5 = \bigg\{}
		\begin{tabular}{l}
		\small{$1$ if $\mid G_r-I_r \mid < 5$} \\
		\small{$0$ otherwise} \\
		\end{tabular}
	\end{aligned}
	\label{eq:hits}
\end{equation}
\vspace{-0.2cm}
\begin{equation}
    \small{\textrm{MRR*} = \frac{1}{N}\sum_{n=1}^{N} \frac{1}{\mid G_r-I_r \mid+1}}
    \label{eq:mrr}
\end{equation}

where $N$ is the number of triples tested, $G_R$ is the ground truth rank, and $I_r$ is the inferred rank. For both these metrics, scores range from 0 to 1 with 1 being the best performance. MRR* is a more complete ranking metric for which the inferred and ground truth ranks must match exactly to get an MRR* 1. Hits@5 gives a more granular look at rankings, but is informative of how often the correct response is within some threshold. We discuss how the ground truth set and ranks are generated for each experiment in Section~\ref{sec:exp_pro}.
    


\section{Experimental Procedure}
\label{sec:exp_pro}

	We evaluated RoboCSE's generalization capability on two scenarios:  inferring the ranks of unseen triples (triple generalization) and accurately predicting the properties and locations of objects in previously unseen environments (environment generalization).  


\subsection{RoboCSE Knowledge Source: AI2Thor}
\label{sec:knowledge_sources}

\begin{table}[b!]
    \vspace{-0.5cm}
	\caption{RoboCSE Knowledge Source}
    \centering      
      \begin{tabular}{@{}lccccc@{}} \toprule
        \multicolumn{6}{c}{AI2Thor: 3 Relation Types, 117 Entities} \\ \midrule
        \multicolumn{6}{c}{Median Count per Environment} \\ \midrule
        \makecell[c]{Env. \\ Type} & \makecell[c]{Loc. \\ Rel.} & \makecell[c]{Mat. \\ Rel.} & \makecell[c]{Aff. \\ Rel.} & Entities & \makecell[c]{Num. \\ Rooms} \\ \midrule
        Bathroom & 28 & 21 & 46 & 18 & 30 \\ 
        Bedroom & 28.5 & 16 & 54.5 & 20 & 30 \\ 
        Kitchen & 59.5 & 51 & 109 & 27 & 30 \\ 
        Livingroom & 22.5 & 8 & 37 & 20 & 30 \\ 
        All & 29.5 & 18.5 & 50 & 20 & 120 \\ \bottomrule
      \end{tabular}
	\label{tbl:thor}
\end{table}

In this work, our semantic reasoning framework targets common sense knowledge for residential service robots.  Knowledge embedded in RoboCSE was mined from a highly realistic simulation environment of household domains, AI2Thor \cite{kolve2017ai2}. AI2Thor offers realistic environments from which instances of semantic triples about affordances and locations of objects can be mined (see Table~\ref{tbl:thor}). Entities include 83 household items (e.g. microwave, toilet, kitchen) and 17 affordances (e.g. pick up, open, turn on). Additionally, we manually extended objects within AI2Thor to model 17 material properties (e.g. wood, fabric, glass), which were assigned probabilistically based on materials encountered in the SUNCG dataset \cite{song2016ssc}. The addition of material properties brought the total number of triples available for training, validation, and testing to over 15K.

   Prior work on multi-relational embedding has shown that inclusion of negative examples in the training data, defined as triples known to be false, leads to improved training performance \cite{nickel2016review}.  To take advantage of this result, we additionally trained on $(9 \times \textrm{number\_of\_true\_triples})$ negative examples for our domain.  
   
   Similar to prior work, we used the closed world assumption to sample negative triples. However, we did not find that using the perturbing method suggested in \cite{liu2017analogical} gave the best results. Instead better results were achieved after filtering perturbed triples to verify the sample was not in the training set. The reason for this empirical phenomenon needs further evaluation.


\subsection{Inferring Unseen Triples: Triple Generalization}
\label{sec:tg}

\begin{figure}
      \centering                    
      \includegraphics[width=0.5\columnwidth]{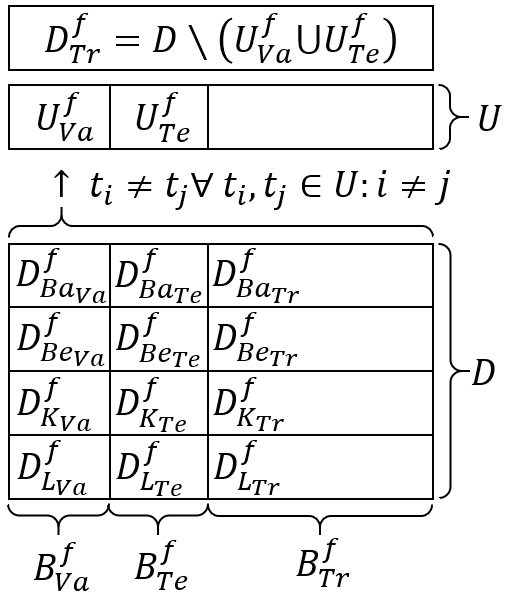}
      \caption[]
      {\small Diagram of triples contained in train, validation, and test sets for each fold (not to scale) during triple generalization (above) and environment generalization (below).} 
      \label{fig:exp_pro}
      \vspace{-0.5cm}
\end{figure}

	Inevitably, an autonomous robot operating in a real-world environment will encounter problems that require answers to queries it was not trained on (e.g. \textit{can mugs be filled?}).  To probe how well RoboCSE can correctly generalize to do triple prediction, triple generalization performance was tested for each algorithm as follows.
    
    
    Five-fold cross-validation was performed to estimate each algorithm's performance. To generate each fold, a set of all unique triples $U$ was generated from the set of all triples in our test case dataset $\mathcal{D}$ where $U \subset \mathcal{D}$ by filtering out repeated triples (i.e. each triple $t \in U$ has the property $t_i \neq t_j \forall t_i,t_j \in U : i \neq j$) (see Figure~\ref{fig:exp_pro}). $U$ was split into five equally sized sets of triples for folds $U^f$ where $f \in \{1,2,3,4,5\}$. $U^f$ was divided in half to create a validation portion $U^f_{Va}$ and test portion $U^f_{Te}$. The training set for each fold $D^f_{Tr}$ was generated from $\mathcal{D}$ by ensuring $\mathcal{D}^f_{Tr} \subset \mathcal{D}$ such that $\mathcal{D}^f_{Tr} \cap (U^f_{Va} \cup U^f_{Te}) = \emptyset$. For each fold, $\mathcal{D}^f_{Tr}$ was trained on while validating on $U^f_{Va}$, and the learned embedding was then tested on $U^f_{Te}$.
    
    The training process follows the same procedure as in \cite{liu2017analogical}. Testing was done by generating three ranks with each triple (i.e. rank $h$ given $(h,r,\line(1,0){5})$, rank $r$ given $(h,\line(1,0){5},t)$, and rank $t$ given $(h,r,\line(1,0){5})$) then comparing them to their respective ground truth ranks. Each triple in the test set was a held-out triple ranked using the full-distribution of triples $\mathcal{D}$. Ground truth ranking was calculated according to the number of observations (i.e. more observations give higher rank). Error metrics similar to those from the relational embedding community (MRR* and Hits@K*) were calculated using the ground truth rank for comparison.


\subsection{Applying Common Sense: Environment Generalization}
\label{sec:eg}

Our second test targets the scenario of deploying a robot equipped with a semantic knowledge base in a new environment, with the goal of evaluating how well the embedded knowledge generalizes to new rooms and the degree to which a robot can use its knowledge to predict object properties or locations in the new setting (e.g. in a new house, where would I likely find a towel?). Environment generalization was tested as follows.	

	Five-fold cross-validation was performed to estimate each algorithm's performance over a test case dataset balanced across environment types (i.e. bathroom, bedroom, kitchen, livingroom). To generate each fold, $\mathcal{D}$ was separated into four sets for each environment type maintaining resolution at environment level (i.e. a single environment with all triples contained is an atomic unit for splitting purposes), $\mathcal{D}_{Ba}$ for bathrooms, $\mathcal{D}_{Be}$ for bedrooms, $\mathcal{D}_{K}$ for kitchens, and $\mathcal{D}_{L}$ for living rooms (see Figure~\ref{fig:exp_pro}). Then each environment type set $\mathcal{D}_{E}$ for $E \in \{Ba,Be,K,L\}$ was split at environment resolution into five equally sized sets of environments for folds $\mathcal{D}_{E}^f$ where $f \in \{1,2,3,4,5\}$. The smaller fraction of each fold of $\mathcal{D}_{E}^f$ was then divided in half to create a validation portion $\mathcal{D}_{E_{Va}}^f$ and test portion $\mathcal{D}_{E_{Te}}^f$, while the larger fraction served as a training set $\mathcal{D}_{E_{Tr}}^f$. Finally, the balanced train $B_{Tr}^f$, validation $B_{Va}^f$, and test $B_{Te}^f$ sets were generated according to: 
	\begin{equation}
	    B_{Tr}^f \! = \! \bigcup_{e \in E} \! \mathcal{D}_{e_{Tr}}^f \quad B_{Va}^f \! = \! \bigcup_{e \in E} \! \mathcal{D}_{e_{Va}}^f \quad B_{Te}^f \! =  \! \bigcup_{e \in E} \! \mathcal{D}_{e_{Te}}^f
	    \label{eq:balance}
	\end{equation}

	The training process followed the same procedure as in \cite{liu2017analogical}.  Testing was done by querying the tested algorithm for triples that come from new environments, which have not been trained on found in each $B_{Te}^f$ for folds $f \in \{1,2,3,4,5\}$. The standard MRR and Hits@K were used to measure the algorithm's performance, allowing us to assess how frequently the robot was correct on the first attempt.


\section{Experimental Results}
\label{sec:exp_res}

	In this section, we report results characterizing the performance of various models trained on AI2Thor data to understand the advantages and limitations of RoboCSE.  Pre-trained Word2Vec embeddings were used in Triple Generalization as a comparable baseline not within the class of multi-relational embeddings. An upper-bound on the performance of logic-based systems was also included in the Triple Generalization experiment to compare with more historically prevalent approaches \cite{tenorth2010knowrob,lemaignan2010oro,suh2007ontology}. For Environment Generalization and Domain Transfer, an instance-based learning baseline that memorizes the training set was used. This controlled baseline gave a clear indicator of how well RoboCSE generalized knowledge beyond what was available in the training set. Lastly, the memory requirements of RoboCSE were compared to a Bayesian Logic Network (BLN) because both account for uncertainty while modeling a graph of knowledge triples unlike Word2Vec or logic-based approaches\footnote{Bayesian Logic Networks and Markov Logic Networks were widely used in previous works but suffer from similar intractability problems \cite{chernovasituated,tenorth2010knowrob,zhu2014reasoning}. Due to memory requirements, the BLN baseline could not be included in all experiments.}.


\subsection{Testing Triple Generalization}
\label{sec:tg_res}

    Triple Generalization was tested to quantify how well RoboCSE could infer missing triples using the learned representation (i.e. infer rank for \textit{fork atLocaion kitchen}, not in the train set).
    
\begin{figure}
    \centering
	\begin{subfigure}[b]{\columnwidth}
        \centering
        \includegraphics[width=\columnwidth]{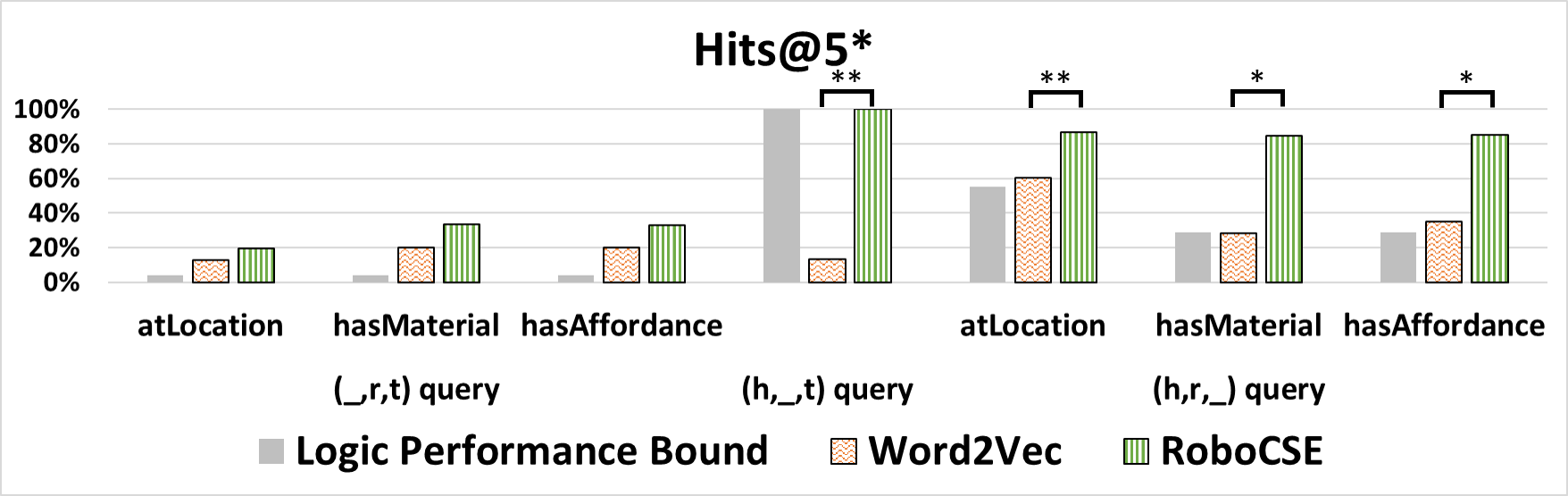}
        \caption[]%
        {{}}    
        \label{fig:tg_hits}
    \end{subfigure}
    \begin{subfigure}[b]{\columnwidth}
        \centering
        \includegraphics[width=\columnwidth]{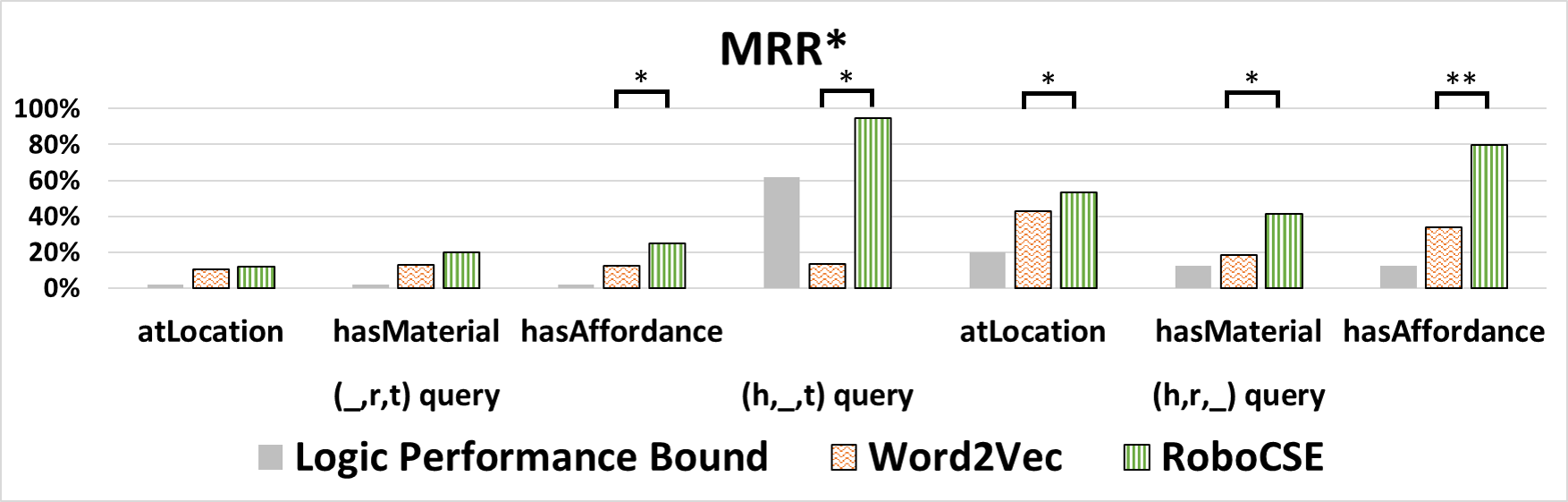}
        \caption[]%
        {{}}
        \label{fig:tg_mrr}
    \end{subfigure}
    \caption[]
    {\small Performance w.r.t. Hits@5* and MRR* metrics for Triple Generalization in AI2Thor}
	\label{fig:tg_bar}\vspace{-0.5cm}
\end{figure}

    Two baselines were used to compare with RoboCSE, Word2Vec and Description Logics (DL) performance upper-bound. The Word2Vec baseline first forms a `comparison' group $C$ of responses from all triples in the training set matching a test query (i.e. given \textit{(\line(1,0){10},atLocation,cabinet)}, group heads from all training triples matching this query). With the Word2Vec embeddings of $C$, the Word2Vec embedding of all candidate responses (i.e. all entities $\notin C$) are ranked using the cosine distance. We estimated the upper-bound of a DL based system at best be able to perform at type-specific chance (e.g. for a total of 17 affordances, guessing the correct affordance to \textit{(mug,hasAffordance,\line(1,0){10})} in the top five hits has a chance $\frac{5}{17}$). This is because DL can determine the type of result that should be returned by a query but cannot infer which entity within a type would be most likely. Therefore the performance could be estimated for each query assuming type-specific chance (see Figure~\ref{fig:tg_hits}). The bar graphs in Figure~\ref{fig:tg_bar} show the performance of each algorithm w.r.t. Hits@5* and MRR* metrics for each relation and query type on the x-axis.

	RoboCSE outperformed all baselines across all metrics at predicting unseen triples, which were statistically significant improvements on $(h,\line(1,0){5},t)$ and $(h,r,\line(1,0){5})$ queries compared using non-parametric 2 group Mann-Whitney U tests. The DL bound performs well for $(h,\line(1,0){5},t)$ queries because DL has explicitly defined types for all entities in a T-Box \cite{baader2008description}, allowing the framework to select the correct relation given a head and tail. The overall implication of these performance improvements is that a robot using RoboCSE to reason about a task could not only infer new knowledge it might not have been trained on to complete a task, but also reason about the confidence in the inferences to return the best result.
    
    All algorithms performed worse at $(\line(1,0){5},r,t)$ queries than other queries, which is prevalent across our experiments. This drop in performance is because selecting the right entity as a head to complete a triple is a more difficult learning problem (a chance of $\frac{1}{74}$) versus selecting the correct affordance, material, or location, which are much fewer in number.
    

\subsection{Testing Environment Generalization}
\label{sec:eg_res}

	Environment Generalization was tested to measure how well RoboCSE could accurately complete triples in new rooms, motivated by real-world application of RoboCSE and the way training/deployment would proceed when a service robot encounters a new environment.
    
    We compared RoboCSE to an instance-based learning baseline that memorizes the training set (i.e. frequency count) and the initial results showed these two methods were comparable. The baseline completed triples by selecting the most observed matching candidate (i.e. given query $(\line(1,0){5},r,t)$ it returned the head most often observed with the matching relation $r$ and tail $t$). We trained each algorithm on 24 rooms of each type available and the results showed the baseline and RoboCSE had closely matching strong performances (whereby performance of each was within 1\% of each other, $>$90\% for $(h,\line(1,0){5},t)$ and $(h,r,\line(1,0){5})$ queries and $>$40\% for $(\line(1,0){5},r,t)$). This was because the default rooms of AI2Thor do not have enough variety between rooms (i.e. algorithms rarely have to generalize to unseen triples).

\begin{figure}[b!]
    \vspace{-0.5cm}
    \centering
	\begin{subfigure}[b]{\columnwidth}
        \centering
        \includegraphics[width=0.75\columnwidth]{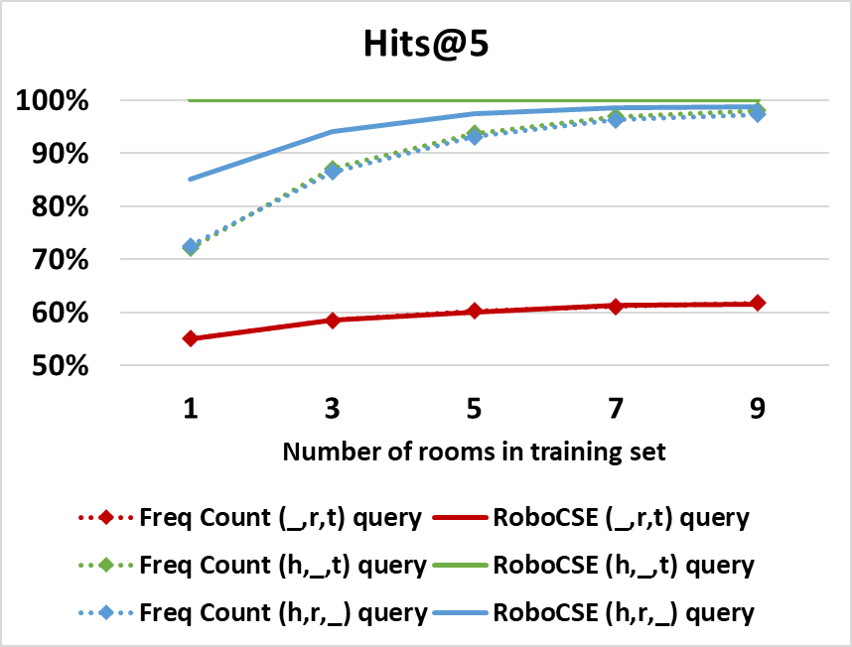}
        \caption[]%
        {{}}    
        \label{fig:eg_hits}
    \end{subfigure}
    \begin{subfigure}[b]{\columnwidth}
        \centering
        \includegraphics[width=0.75\columnwidth]{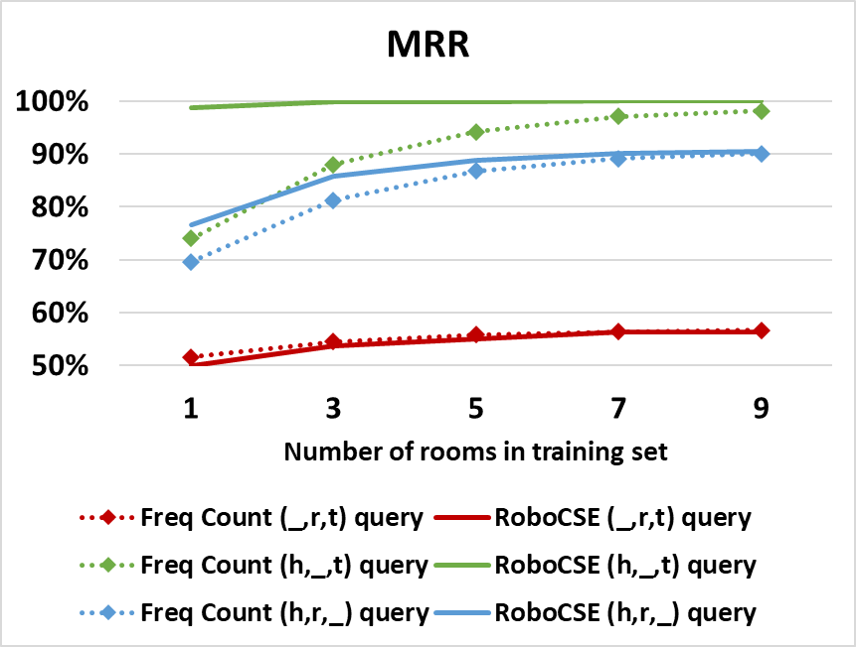}
        \caption[]%
        {{}}
        \label{fig:eg_mrr}
    \end{subfigure}
    \caption[]
    {\small Performance Trends w.r.t. Hits@5* and MRR* Metrics for Environment Generalization in AI2Thor}
	\label{fig:eg_trend}
\end{figure}
    
    However, reducing the number of rooms reveals RoboCSE's ability to learn from the interactions of triples and generalize to the best performance faster than the baseline (see Figure~\ref{fig:eg_trend}). Lines in this plot were generated by averaging across relations for each query type at varying numbers of rooms in the training set. The trend of RoboCSE generalizing to new rooms faster than the baseline was most pronounced with the fewest number of rooms in the training set (i.e. 1) but continued up to about 9 rooms as shown in the line plots. We saw this most pronounced for the $(h,\line(1,0){5},t)$ and $(h,r,\line(1,0){5})$ queries on both metrics. This showed from an application's perspective how a robot bootstrapped with RoboCSE can learn general structures from individual instances to perform better in new environments and require less training data.

\subsection{Domain Transfer: Testing on MatterPort3D}
\label{sec:transfer}

    The learned embeddings from AI2Thor were tested on MatterPort3D (MP3D) to measure how well RoboCSE transfers to \textit{envrionments from} real-world domains. While MP3D does not contain all the object properties we included in AI2Thor (no affordances or materials), it does contain triples about object locations for over 500 real-world environments.

\begin{figure}
    \centering
	\begin{subfigure}[b]{\columnwidth}
        \centering
        \includegraphics[width=\columnwidth]{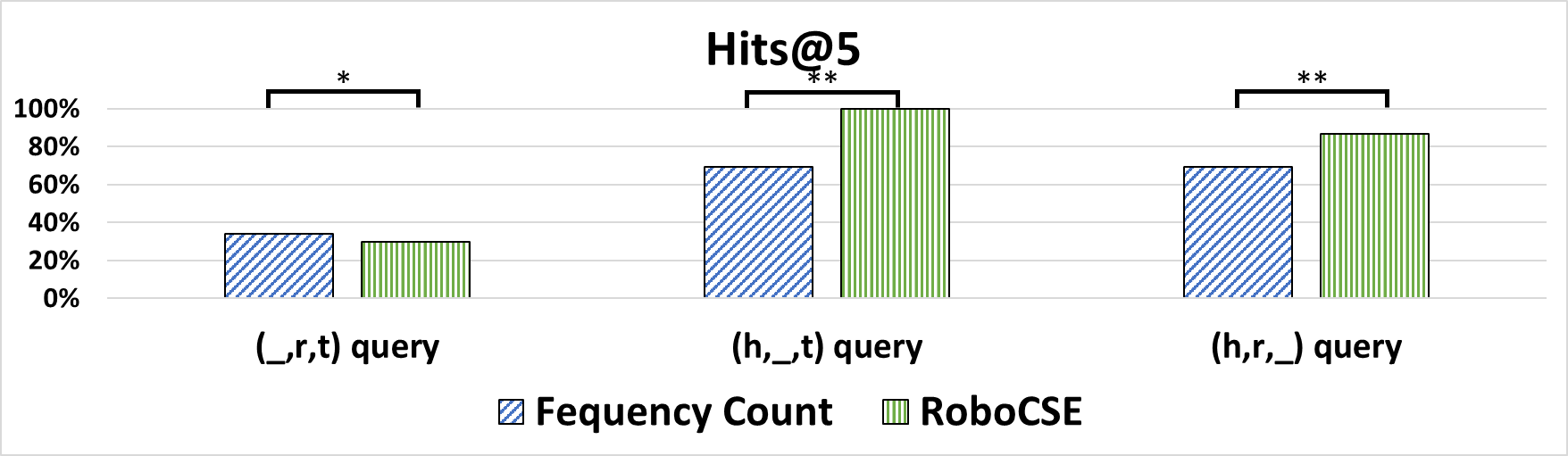}
        \caption[]%
        {{}}    
        \label{fig:mp3d_hits}
    \end{subfigure}
    \begin{subfigure}[b]{\columnwidth}
        \centering
        \includegraphics[width=\columnwidth]{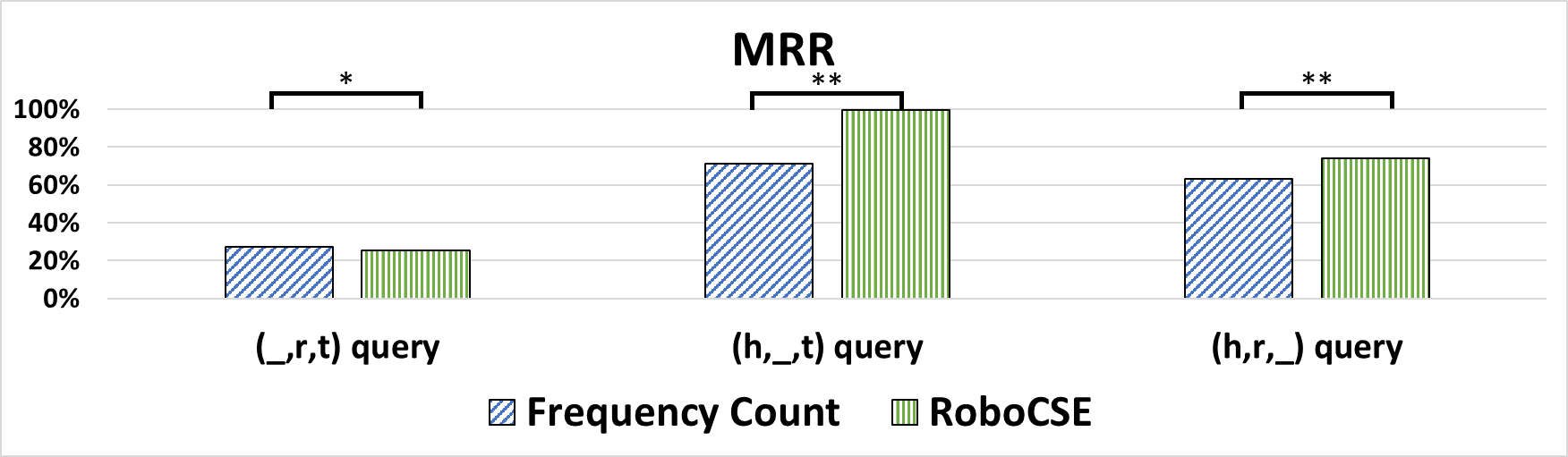}
        \caption[]%
        {{}}
        \label{fig:mp3d_mrr}
    \end{subfigure}
    \caption[]
    {\small Performance w.r.t. Hits@5 and MRR metrics for Domain Transfer to MatterPort3D}
	\label{fig:transfer}
    \vspace{-0.5cm}
\end{figure}

    The results from domain transfer showed that RoboCSE generalized to MP3D better than our instance-based learning baseline that memorizes the training set (i.e. frequency count), effectively inferring new triples not present in the training data. Training and validation for domain transfer closely followed the Environment Generalization procedure (see Section~\ref{sec:eg}) but only for \textit{atLocation} relations.  During testing, the models learned from all rooms in AI2Thor were used to answer queries about all rooms in MP3D. The bar graphs in Figure~\ref{fig:transfer} show that the semantics learned in AI2Thor can be directly applied to MatterPort3D, evident in the high performance of both algorithms. Furthermore, inference in RoboCSE successfully generalized beyond training data to accurately infer more queries indicated by the statistically significant higher scores RoboCSE gets on $(h,\line(1,0){5},t)$ and $(h,r,\line(1,0){5})$ queries compared using non-parametric 2 group Mann-Whitney U tests. In short, this shows that bootstrapping a robot with semantics learned in simulation using RoboCSE can be applied to data from real world environments.


\subsection{Analyzing Memory Requirements}
\label{sec:scale_res}

	We analyzed the memory requirements of RoboCSE and BLNs \cite{jain2009bayesian} to compare the scalability of each.

    To analyze memory requirements, all unique triples from AI2Thor were extracted (352) and modeled in a BLN using a standard package (ProbCog \cite{jain2010soft}). The resulting BLN required 9 orders of magnitude more memory than RoboCSE (i.e. 100 TB vs. 96 KB). Although BLNs have been used to model semantic knowledge within robot systems to do accurate probabilistic reasoning \cite{tenorth2010knowrob,chernovasituated}, maintaining conditional-probability tables in BLNs can be intractable due to the rapid increase of node in-degree (i.e. number of parents) and therefore table size, for densely connected networks.
 	
    RoboCSE's drastic memory reduction was possible because its space complexity scales linearly with the number of entity or relation types and RoboCSE's space complexity does not directly depend on node in-degree. RoboCSE requires (\textit{number\_of\_entities}$+$\textit{number\_of\_relations})$\times d \times 8$ bytes of memory, where $d$ is the vector space dimensionality. While this is a considerable improvement in space complexity, RoboCSE cannot represent the joint distribution or true probabilities as a BLN can. Instead, the distances measured using a scoring function between the queried transformation and results are interpreted as confidence (see Section~\ref{sec:details}). Furthermore, only the subset of the triple in the query can be used as `evidence' to condition on (e.g. the best $h_i$ are selected conditioned on an $(\line(1,0){5},r,t)$ query).

\vspace{-0.1cm}
\section{Discussion \& Conclusion}
\label{sec:discussion}

	In this work we approached the problem of semantically representing a robot's world via $(h,r,t)$ triples in a manner that supports generalization, accounts for uncertainty, and is memory-efficient. We presented RoboCSE, a novel framework for robot semantic knowledge that leverages mutli-relatonal embeddings.
	
	From our experiments two benefits have emerged from the use of multi-relational embeddings in RoboCSE: (1) the generalizations learned outperformed Word2Vec at prediction, being robust to significant reductions in training data and domain transfer and (2) RoboCSE used orders of magnitude less memory to represent projections of graphs than representations of the same graph with BLNs. The collectively distinct set of benefits multi-relational embeddings have to offer could be taken advantage of to further progress for robots performing semantic reasoning robustly in semantically rich environments.

    However, leveraging multi-relational embeddings has its limitations. As previously mentioned, answering $(\line(1,0){10},r,t)$ queries is particularly difficult. This query is useful for robots reasoning to plan tasks (i.e. which head satisfies \textit{(\line(1,0){10},hasAffordance,fill)}). Secondly, conditioning is very limited compared to a BLN. This leads to the same responses in different environments. Lastly, realistic systems in long-term deployments need the ability of incremental learning, enabling online adaptations as new knowledge arrives, which is not possible in this RoboCSE formulation.
    

\vspace{-0.1cm}
\section*{ACKNOWLEDGMENT}
{This work is supported in part by NSF IIS 1564080, NSF GRFP DGE-1650044, and ONR N000141612835. Zsolt Kira was partially supported by NRI/NSF grant \#IIS-1426998. Any opinions, findings, and conclusions or recommendations expressed in this material are those of the author(s) and do not necessarily reflect the views of the supporters.}
\clearpage



\addtolength{\textheight}{-12cm}   

\bibliographystyle{IEEEtran}
\bibliography{biblio}

\end{document}